\documentclass[letterpaper, 10 pt, conference]{ieeeconf}  

\IEEEoverridecommandlockouts                              

\usepackage{amsmath} 
\usepackage{amssymb}  
\usepackage{bm}
\usepackage{graphicx}
\usepackage[utf8]{inputenc}
\usepackage{epstopdf}
\usepackage{siunitx}
\usepackage{booktabs}
\usepackage{url}
\usepackage{multirow}
\usepackage{grffile}
\usepackage{adjustbox}
\usepackage{tikz}

\newcolumntype{R}[2]{%
    >{\adjustbox{angle=#1,lap=\width-(#2)}\bgroup}%
    l%
    <{\egroup}%
}
\newcommand*\rot{\multicolumn{1}{R{20}{1em}}}  

\title{\LARGE \bf
When Geometry is not Enough: \\
Using Reflector Markers in Lidar SLAM
}

\author{Gerhard Kurz\textsuperscript{\textsection}$^{1}$, Sebastian A. Scherer\textsuperscript{\textsection}$^{1}$, Peter Biber$^{1}$ and David Fleer$^{2}$
\thanks{$^{1}$Gerhard Kurz, Sebastian A. Scherer, Peter Biber are with Robert Bosch Corporate Research, Germany. {\tt\small \{gerhard.kurz2, sebastian.scherer2, peter.biber\}@de.bosch.com}}%
\thanks{$^{2}$David Fleer is with Bosch Rexroth AG, Germany. {\tt\small david.fleer@de.bosch.com}}%
}

\newcommand\copyrighttext{%
	\footnotesize \textcopyright 2022 IEEE. Personal use of this material is permitted.
	Permission from IEEE must be obtained for all other uses, in any current or future
	media, including reprinting/republishing this material for advertising or promotional
	purposes, creating new collective works, for resale or redistribution to servers or
	lists, or reuse of any copyrighted component of this work in other works.
}
\newcommand\copyrightnotice{%
	\begin{tikzpicture}[remember picture,overlay]
	\node[anchor=south,yshift=10pt] at (current page.south) {\fbox{\parbox{\dimexpr\textwidth-\fboxsep-\fboxrule\relax}{\copyrighttext}}};
	\end{tikzpicture}%
}

\begin{document}

\maketitle
\copyrightnotice
\begingroup\renewcommand\thefootnote{\textsection}
\footnotetext{Equal contribution}
\endgroup

\thispagestyle{empty}
\pagestyle{empty}

\begin{abstract}
Lidar-based SLAM systems perform well in a wide range of circumstances by relying on the geometry of the environment.
However, even mature and reliable approaches struggle when the environment contains structureless areas such as long hallways.
To allow the use of lidar-based SLAM in such environments, we propose to add reflector markers in specific locations that would otherwise be difficult.
We present an algorithm to reliably detect these markers and two approaches to fuse the detected markers with geometry-based scan matching.
The performance of the proposed methods is demonstrated on real-world datasets from several industrial environments.
\end{abstract}

\section{INTRODUCTION}
In recent years, lidar-based SLAM systems have become widespread in a variety of robotics applications ranging from intralogistics robots and cleaning robots to service robots and autonomous vehicles.
While they perform well in many different indoor and outdoor environments, there are still cases where they struggle, such as structureless environments (e.g., corridors), highly dynamic environments and on uneven ground (e.g., on ramps).
These issues are especially apparent when 2D lidar sensors are used, but can also affect 3D lidar SLAM, although usually to a lesser degree.

To address these issues, we propose to place unobtrusive reflector markers in the particular areas where traditional lidar-based SLAM performs inadequately and to use these markers inside the lidar odometry of our SLAM system.
This, way, we artificially create structure in the environment to be used by the SLAM algorithm and enable the use of lidar SLAM in environments that would be unsuitable otherwise.
Since we only need to add a limited number of artificial markers in selected difficult areas, we mostly preserve the advantages of infrastructureless lidar-based SLAM methods, where the environment does not need to be modified to enable the use of SLAM.

We use retroreflective tape as reflector markers, since it can easily be placed on existing walls or objects in the environment and does not introduce new obstacles in potentially confined environments.
Also, unlike some other marker-based methods, we do not assume that the marker locations are known in advance or chosen with particular care, as long as difficult areas are covered sufficiently.

\begin{figure}
	\centering
	\includegraphics[width=0.99\linewidth]{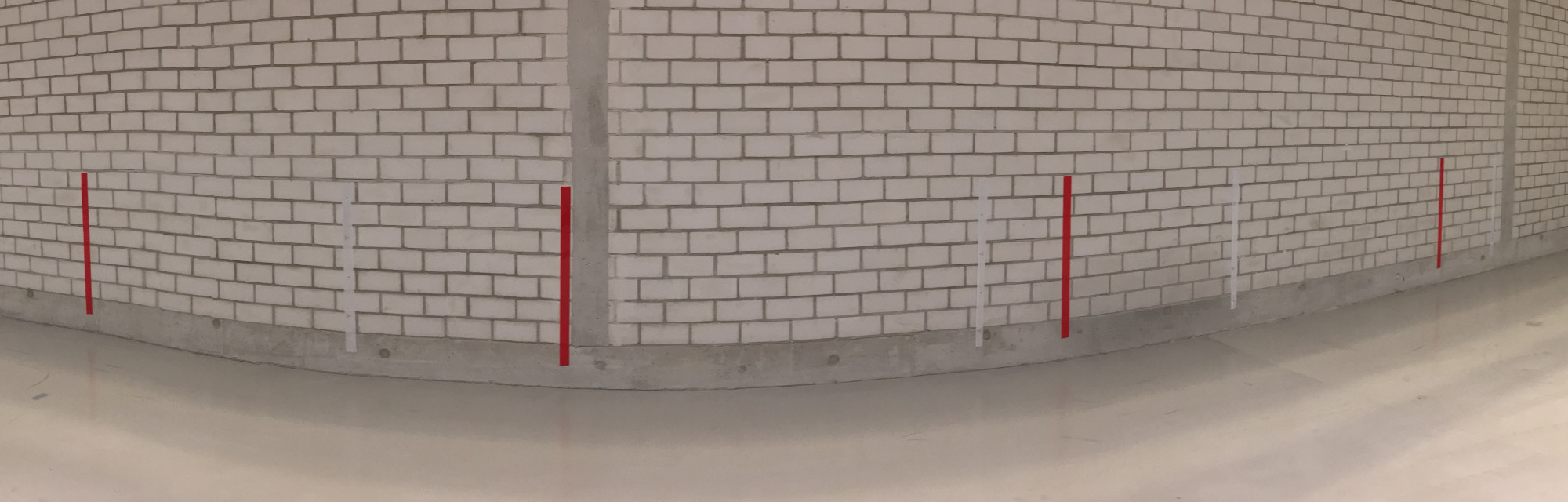}
	\caption{Example of a long, featureless corridor marked with retroreflective tape.}
    \label{fig:reflectorsCorridor}
\end{figure}

Specifically, the contributions of this paper are:
\begin{enumerate}
	\item An algorithm to reliably detect reflector markers in laser scans that can easily be adapted to different lidar sensors.
	\item An approach to use reflector markers in a scan matcher by integrating them as a separate semantic layer within the map against which scans are matched.
	\item A second approach to use reflector markers in a scan matcher by explicitly tracking markers and adding a tracking-based term to the cost function.
	\item Integration of reflector markers into an existing scan-matching-based lidar SLAM system.
	\item Thorough evaluation of the proposed approaches in multiple industrial environments.
\end{enumerate}

\section{RELATED WORK}
Lidar-based localization and mapping based on reflector markers alone has been around for decades \cite{Guivant2000}.
The first generation of commercially-available lidar localization systems for mobile robots also relied on markers, which had to be carefully placed and surveyed in the desired working environment.
Using markers is attractive because it allows reliable, accurate localization and can easily be implemented even on computationally-constrained embedded devices.
For this reason, they are still used for vehicle localization with markers on guardrails~\cite{Ghallabi2019}, for automated robots with many tube-shaped reflectors placed in the environment~\cite{Wang2021}, or for localization in rough terrain~\cite{Davis2019}.
However, setting up the reflector markers can be tedious. Relying on markers also constrains the area of operation to the environment where markers were placed.

Being able to perform lidar localization and mapping without artificial landmarks by matching a whole laser scan to reference scans is therefore a big advantage.
Almost all lidar-based localization and mapping systems today rely on scan matching (either ICP~\cite{Besl1992}, NDT~\cite{Biber2003}, or correlative scan matching~\cite{Olson2009}) to find the relative pose between scans.
They then use a graph-based back-end to optimize the global estimate~\cite{Grisetti2010a} based on these poses.

While there are systems that can combine both natural features and artificial markers for visual SLAM (e.g.~\cite{MunozSalinas2020}), we did not find comparable lidar-based SLAM systems described in the literature.
The systems closest to what we will later call our NDT layer approach are \cite{Zaganidis2018} and \cite{Zaganidis2017}, which are in turn based on \cite{Nuechter2005}, since they also treat points of different semantic classes separately during scan matching.

\section{SYSTEM ARCHITECTURE}
\begin{figure}
	\centering
	\includegraphics[width=0.99\linewidth]{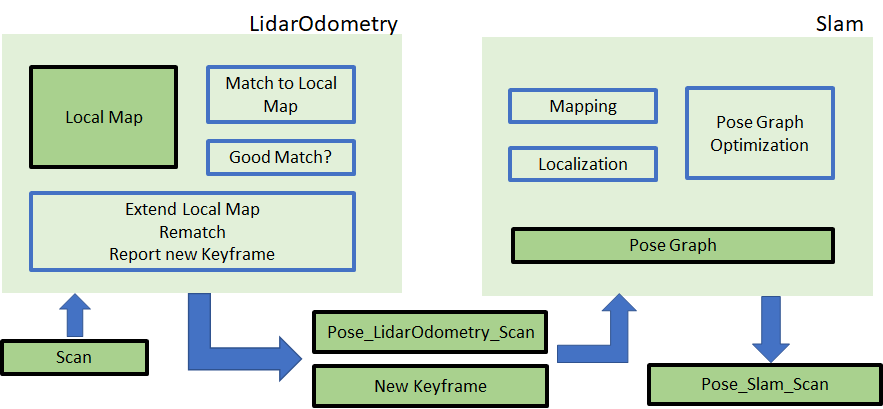}
	\caption{Rough system architecture. The lower part shows the data flow. A scan is first processed by the lidar odometry, the output is the pose of the scan in lidar odometry frame and potentially a new keyframe. SLAM can either run in mapping or localization mode, and outputs the pose in map frame. See main text for details. The focus of the paper is the lidar odometry part.}
    \label{fig:sysar}
\end{figure}

Our general system architecture is shown in Fig.~\ref{fig:sysar}. We first give an overview and then go into some details. A conclusion of this section is the design decision to integrate reflector marker support within the lidar odometry, but not in the mapping or localization specific modules. The keyframe selection mechanism described here is also relevant for the further sections.

The system expects laser scans as inputs and outputs a pose for each laser scan. It can run in mapping or localization mode. In mapping mode, a map frame is instantiated at the start pose; in localization mode a global map and an initial pose are needed as additional inputs. The pose is then output with respect to the frame of the map. Initial poses can be provided by the user or by a global localization procedure (not described here).

In both modes, the input laser scan is first processed by the lidar odometry. The lidar odometry takes scans as inputs and outputs poses within the lidar odometry frame. This lidar odometry frame starts at zero for the first scan. In addition to the pose, the lidar odometry also reports new keyframes that are used for localization or mapping.

Our lidar odometry maintains a local map and the current pose. When a new scan arrives, it is matched to the local map using NDT scan matching \cite{Biber2003}. After matching, the result is examined to check whether it is a good match. If this is the case, the matching result is directly output as the lidar odometry pose. Otherwise, the local map is extended and the current scan is re-matched to the new local map.

Whenever the local map is extended, the added scan is declared a new keyframe. This new keyframe is then added to a pose graph. It is connected to the previous keyframe by a lidar odometry edge using the respective scan matching result.
In mapping mode, we then search for loop closures and add binary edges for successful matches; in localization mode, we match scans to the global map and add unary edges for successful matches.
Here, the pose graph chain is limited to a maximum number of vertices (e.g., 50).
In both modes we handle only keyframes.

After adding the lidar odometry edge and potentially loop closure or map matching edges, the pose graph is optimized using g2o~\cite{Kuemmerle2011}. The resulting pose of the latest keyframe is then used to update the transformation between lidar odometry and map frames.
The pose can then always be output in the map frame, both for keyframe and non-keyframe scans.

Going into the details, the initial pose for matching the current scan to the local map can be derived from various sources. If a good wheel odometry or IMU is available, it can serve both as an initial estimate and as a prior for scan matching \cite{Andreasson2017EgoMotionUncertainty}. Using these additional sensors, we can easily handle, e.g., motion through a featureless corridor. However, in the following, we consider the case where we have no such additional sensors. The lidar sensor may also have a short range or limited field of view, for example when retrofitting existing vehicles like forklifts. In this scenario, we can only use a constant velocity assumption to derive the initial pose estimate. However, it would be unreliable to depend solely on such an assumption when determining the prior for scan matching. If the matching fails, we could indicate this and handle it in later stages of our pipeline. Nevertheless, from a system architecture point of view, it is easier if later stages can always rely on the lidar odometry output. This is one reason to include reflector marker handling in the lidar odometry.

Another detail is the keyframe selection. Here, we check the scan matching result for the matching score and for the overlap with the local map. This overlap is the ratio of points in the input scan that can be matched to valid NDT cells in the local map, relative to the total number of points in the input scan. Additionally, we also limit the maximum translation and rotation distances.
The overlap check ensures that the local map is extended if there is something new to be mapped, even if the old local map would be completely sufficient for exact scan matching.
Parameter values are chosen in a conservative way, so that a match that passes the check is nearly always correct. When the check fails for the current scan, we add the last scan that passed the check to the local map. This scan will then become the new current keyframe. This ensures consistency of the local map. The local map is maintained as a buffer of the past $n$ keyframes, so when we add a new keyframe we drop the oldest one if the buffer is already full. In rare cases, the first scan matching after adding a new keyframe (i.e., the re-matching of the current scan) may fail. In this case, we output the initial estimate of the matching attempt with a high covariance.

One might be tempted to also let reflector markers help in localization and mapping stages, i.e., for loop closure finding and map matching. There are two reasons not to do this: First, there is the obvious danger that markers are relocated, or maybe just moved a little bit. The local mapping in lidar odometry is unaffected by this, as it always starts from scratch in each run. For localization however, such movements might lead to wrong map matches. A check for changed marker positions would require a complex and potentially not robust map update handling. The second reason is that we would also like to be able to add markers to the real scene after recording the map, with minimal or even no need for user interaction with the SLAM system. In practice, mapping is done only once and often succeeds even in the addressed corner cases. Only later in localization mode do failures occur, usually very sporadically. Integrating markers into our lidar odometry solves these issues without a need for re-recording and distributing a map.

In the next section we will describe detection of reflector markers, and then give two approaches how to incorporate these detection results within the architecture described here.

\section{DETECTION}
\label{sec:detection}

\renewcommand{\vec}{\bm}
\newcommand{\vecp}{\vec{p}}
\newcommand{\vect}{\vec{t}}
\newcommand{\vecd}{\vec{d}}

\begin{figure}
	\centering
	\includegraphics[width=.85\linewidth]{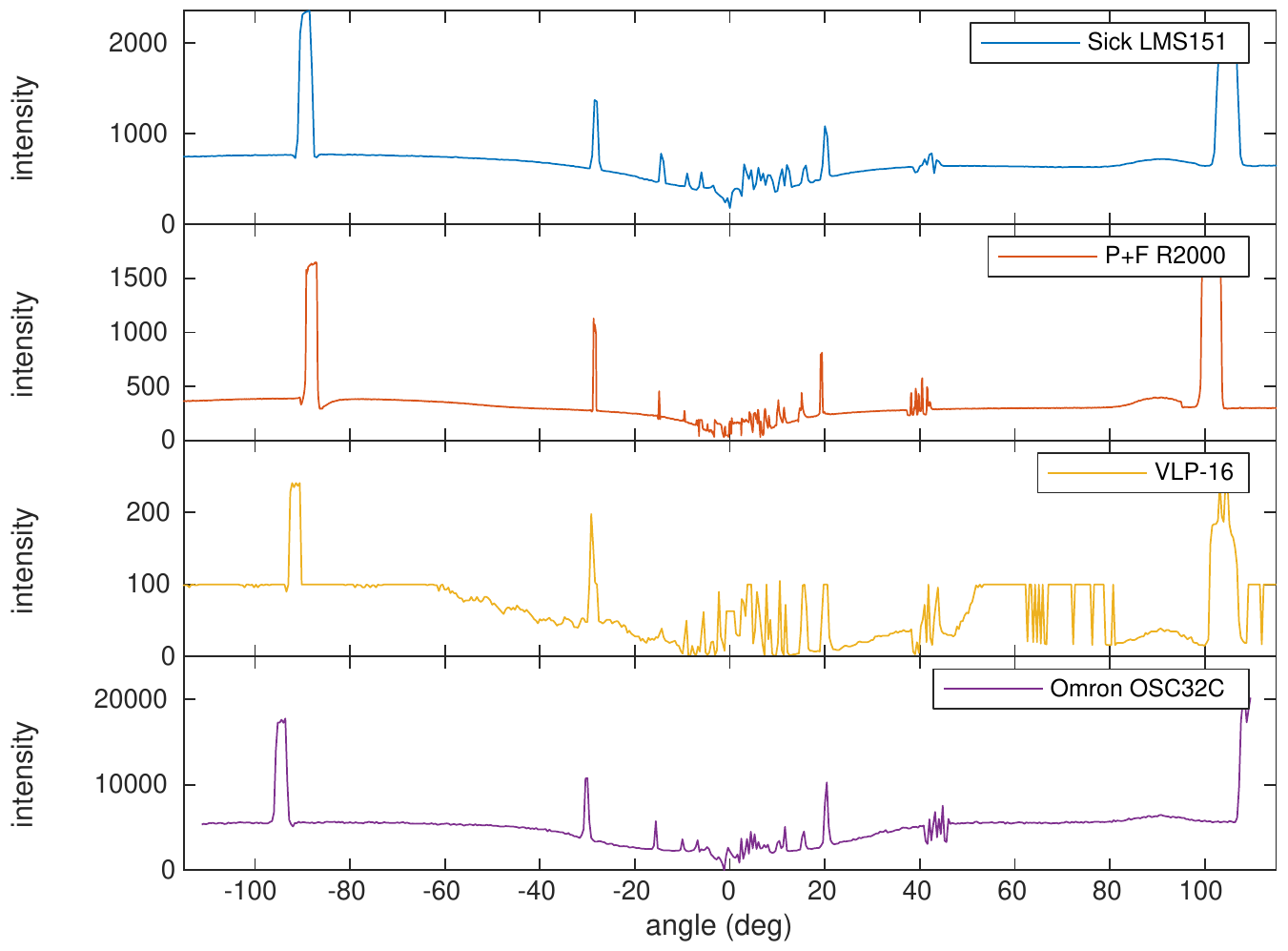}
	\caption{Intensities of different lidars observing the same scene. Note that the $y$-axis is  different as each lidar scales intensities completely differently.
	         Small differences in the shape of the signal are present due to the different mounting positions of the lidars (see Fig.~\ref{fig:Waegele}).}
	\label{fig:compare_lidars}
\end{figure}

Reliably detecting reflector markers in laser scans is a nontrivial task, as it heavily depends on the particular lidar sensor and the environment.
In general, most modern lidar sensors provide not only a range measurement but also an intensity value for each measured point.
While the range is usually measured in meters or millimeters, the intensity value is typically given in a lidar-specific unit, i.e., different sensor models will report widely different ranges of values (see Fig.~\ref{fig:compare_lidars}).

\begin{figure}
	\centering
	\includegraphics[width=.8\linewidth]{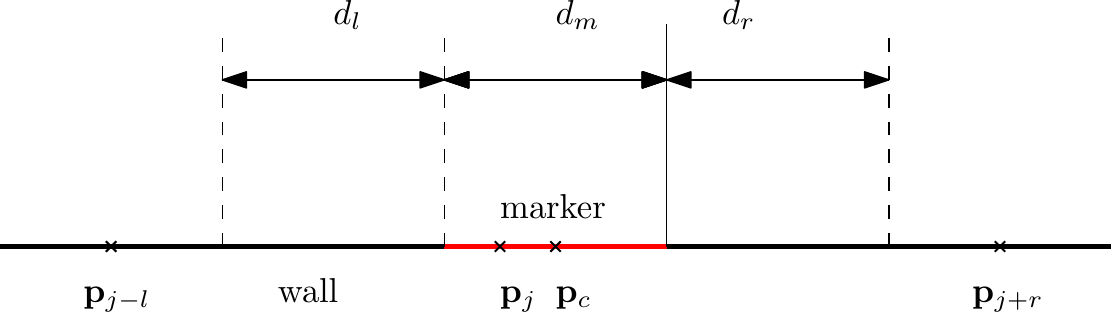}
	\caption{Reflector marker on a wall as seen from the top.}
	\label{fig:wallreflectormarker}
\end{figure}

A simple approach to detect reflector markers would be to simply introduce a (lidar-specific) threshold (see Table~\ref{tab:lidars}) for the intensity and consider all points above this threshold as marker points.
However, this achieves fairly poor reliability, since there might be other non-marker objects in the environment that reflect laser light at a high intensity.
These include reflective surfaces such as metal, safety vests worn by workers, or safety reflectors on vehicles.
In particular, specular reflections can occur even on only moderately reflective surfaces when a lidar ray hits these at a right angle, which is especially likely for round objects like legs of tables. 
To increase robustness in these cases, we propose to place markers of a known width (e.g., $d_m = \SI{5}{\centi\meter}$) on a planar wall, which extends further on both sides of the marker (e.g., $d_l = d_r = \SI{5}{\centi\meter}$ on the left and right).
In the following, we introduce a detector that can reliably detect these kinds of markers (see Fig.~\ref{fig:wallreflectormarker}) even in difficult environments.

The proposed detector operates as follows: First, we consider every point $\vecp_j$ in the laser scan as a possible marker candidate and perform a variety of checks. Within one scan, multiple laser beams may hit the same marker, resulting in multiple marker points being detected. Later, we therefore remove potential duplicate markers based on their Euclidean distance.

\newcommand{\paramminintensity}{i_\text{min}}
\newcommand{\paramminrange}{r_\text{min}}
\newcommand{\parammaxrange}{r_\text{max}}
\newcommand{\paramwindowsize}{w}
\newcommand{\paramminwalllengthm}{l_\text{min}}
\newcommand{\paramminwalllengthp}{p_\text{min}}
\newcommand{\paramregression}{e}
\newcommand{\parampointdelta}{p_d}
\newcommand{\paramdeltaintensityfactor}{c_i}

\begin{table}
	\centering
	\begin{tabular}{ccccccc}
		\toprule
		$\paramminrange$ & $\parammaxrange$ & $\paramwindowsize$ & $\paramminwalllengthm$ &
		$\paramminwalllengthp$ & $\paramregression$ & $\paramdeltaintensityfactor$ \\
		\midrule
		\SI{0.5}{\meter} & \SI{6}{\meter} & \SI{0.15}{\meter} & \SI{0.15}{\meter} & 5 & \SI{0.01}{\square\meter} & 0.333 \\
		\bottomrule
	\end{tabular}
	\caption{Parameter values used in our experiments.}
	\label{tab:parameters}
\end{table}

\begin{table}
	\centering
	\begin{tabular}{lccccc} 
		\toprule
		lidar & freq & FOV & angular res. & $\paramminintensity$ & $\parampointdelta$\\
		\midrule
		SICK LMS151 & \SI{50}{\hertz} & \SI{270}{\degree}  & \SI{0.5}{\degree} & 1000 & 1 \\
		P\&F R2000 & \SI{50}{\hertz}   & \SI{360}{\degree} & \SI{0.1}{\degree} & 500 & 2\\
		Omron OS32C & \SI{13}{\hertz}  & \SI{270}{\degree} & \SI{0.4}{\degree} & 8000 & 1\\
		\bottomrule
	\end{tabular}
	\caption{Specifications of the different lidars in our experiments and the lidar intensity threshold $\paramminintensity$ we chose.}
	\label{tab:lidars}
\end{table}

First of all, the lidar point's intensity needs to be above a predefined threshold $\paramminintensity$, which, of course, is lidar-specific (see Fig.~\ref{fig:compare_lidars}, Table~\ref{tab:lidars}). While this is not a sufficient criterion, it already allows us to eliminate a great percentage of candidate points. We also skip points that are closer than $\paramminrange$ to the lidar, because they may belong to the robot/vehicle. Additionally, we skip points further away than $\parammaxrange$: At long ranges, too few lidar beams will hit the marker, resulting in too few high-intensity points for a reliable detection. Then, we try to identify the wall segment $(\vecp_{j-l}, \dots, \vecp_j, \dots, \vecp_{j+r}$) the point is located on by adding neighboring points from the laser scan on either side until we observe that the Euclidean distance from $p_j$ is larger than a given window size $\paramwindowsize$. This also means that the wall segment will automatically stop at a depth jump. To make sure the detected wall segment is large enough, we check that its length in meters $\lVert \vecp_{j-l} - \vecp_{j+r} \rVert \geq \paramminwalllengthm$ and that the number of points $l+r+1 \geq \paramminwalllengthp$ on the wall segment is sufficient.

To determine the exact borders of the marker, we consider the largest intensity differences inside the wall segment. The greatest upward jump in intensity is assumed to be the beginning of the marker and the greatest downward jump is assumed to be the end. We only consider jumps as valid if the magnitude of each jump is at least $\paramdeltaintensityfactor \cdot \paramminintensity$. We verify that the start of the marker is before $\vecp_j$ and the end is after $\vecp_j$. Now, we determine the center of the marker $\vecp_c$, which may not be exactly identical to $\vecp_j$, by computing the center of mass of all points on the marker. Alternatively, the point with the highest intensity within the marker could also be used as the center point.

Then, we use least-squares regression to fit a line to the marker points. To ensure that the wall is straight, the mean squared error of the regression needs to be below a threshold $\paramregression$. Based on this line, we also compute the normal of the candidate marker. If specular reflections are a problem in the given environment, markers where the normal is pointing exactly towards the lidar can be discarded since they may be the result of specular reflections. However, this comes at the cost of also missing valid markers that just happen to have a normal pointing towards the lidar. Also markers seen at a flat angle can be discarded because their detection tends to be unreliable. Our current implementation discards only points with a flat angle of $\theta> \SI{80}{\degree}$.

\begin{figure}
	\centering
	\includegraphics[width=.9\linewidth]{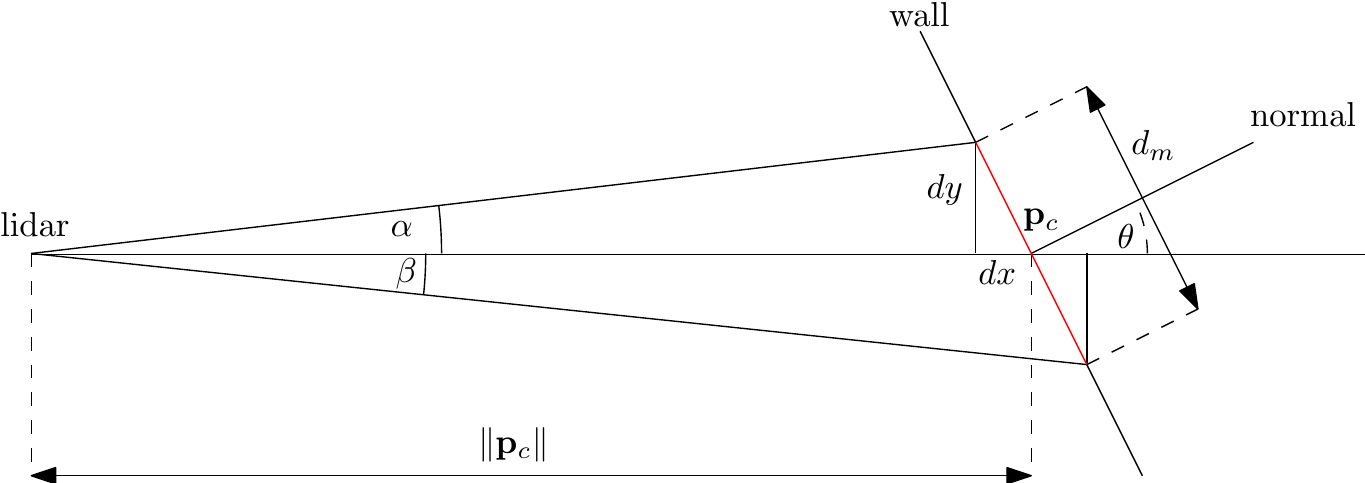}
	\caption{Lidar sensor observing a marker on a wall as seen from the top.}
	\label{fig:lidarmarkerpoints}
\end{figure}

Based on the normal and the distance of $p_c$ from the lidar origin, we can derive the expected size of the marker in terms of lidar points. For this purpose, we compute the angular range $\alpha + \beta$ of lidar rays that hit the marker as depicted in Fig.~\ref{fig:lidarmarkerpoints}. It holds that
\begin{align*}
	dx &= \frac{d_m}{2} \sin \theta \ , \quad
	dy = \frac{d_m}{2} \cos \theta \\
	\alpha &= \mathrm{atan2 }(dy, \Vert \vecp_c \Vert-dx) \ , \quad
	\beta = \mathrm{atan2 }(dy, \Vert \vecp_c \Vert+dx) \ . 
\end{align*}
Using the known angular resolution of the lidar (see Table~\ref{tab:lidars}), we can then compute how many lidar points are expected to hit a marker of size $d_m$ and compare this to the actual number of points on the marker.
If the absolute difference is below a threshold $\parampointdelta$, we assume that the marker is valid and add it to our list of detected markers.
Beyond that, we observed that the number of points detected on a marker is usually $2$ points greater than predicted.
This is caused by the divergence of the laser beam, which means that one beam off to each side of the marker will still partially hit said marker.
Therefore, we always add $2$ to the expected number of points.

\begin{figure}
	\centering
	\includegraphics[width=0.9\linewidth,trim={0 25mm 0 2cm},clip]{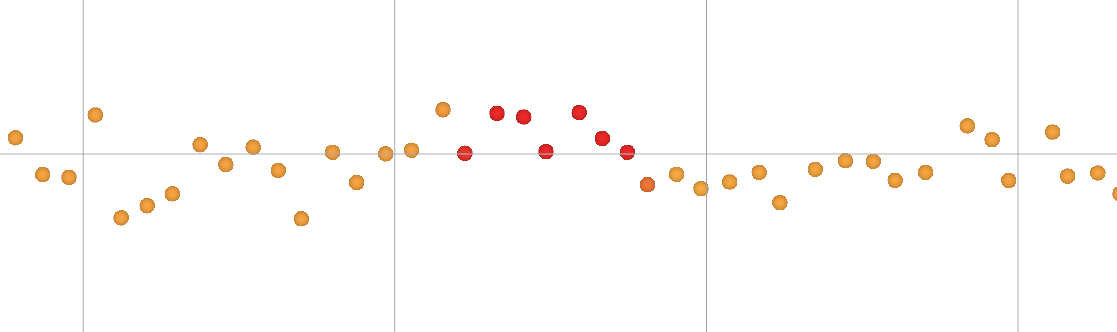}
	\caption{A \SI{5}{\centi\meter} wide marker placed on a wall as seen by a SICK LMS151 lidar with a spatial resolution of \SI{0.5}{\radian} at a distance of approximately \SI{1}{\meter}. The intensity is color-coded: Yellow indicates a low intensity and red indicates a high intensity. The grid cells have a size of \SI{10}{\centi\meter}.}
\end{figure}

The parameters used for the detector are given in Table~\ref{tab:parameters}.

\section{NDT LAYER APPROACH}
Our first approach integrates reflector markers by dividing all points of the laser scan into two sets: Regular points and points on reflector markers.
The overall idea is to perform scan matching for both point sets in parallel, minimizing a linear combination of both cost functions.
The reference point cloud, against which the scan should be matched, is also divided into regular and marker points.
From the resulting two reference point clouds, we build separate NDT representations for these two categories.
During scan matching, we associate each point only to its corresponding NDT representation and optimize cost functions for both point sets in parallel.
This is similar to \cite{Nuechter2005,Zaganidis2017}, in which only points of the same semantic class are associated for scan matching.

\subsection{NDT Representation}
\begin{figure}[ht]
	\centering
    \includegraphics[width=0.48\textwidth,trim={0 5mm 0 3cm},clip]{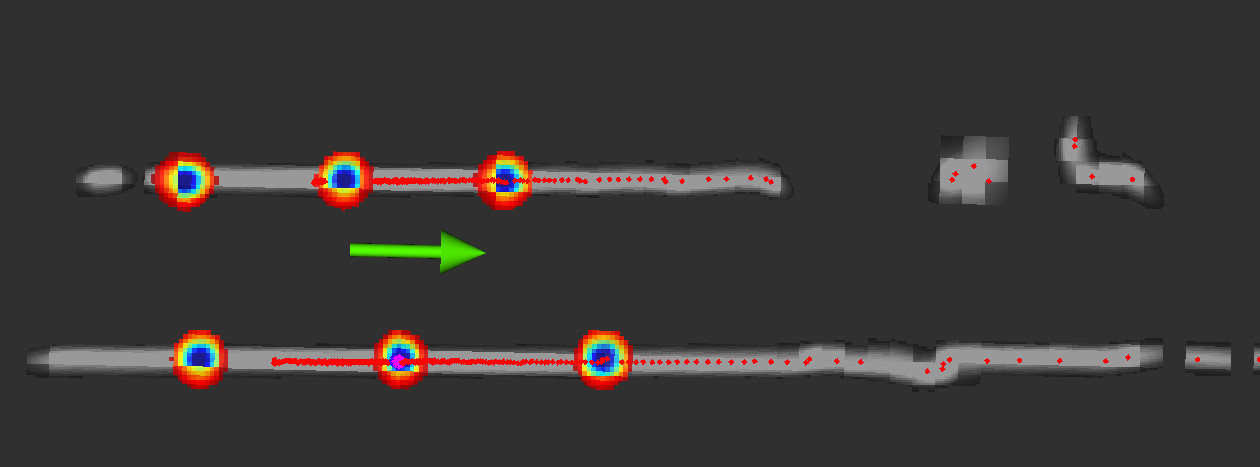}
	\caption{Example of NDT representations of regular points (gray) and marker points (colored).}
	\label{fig:NdtLayers}
\end{figure}

For each reference point cloud, we construct separate NDT representations for regular and marker points.
In its original definition, it requires at least three points to fill a 2D NDT cell with mean and covariance matrix.
Marker points, however, may be relatively sparse, as only a few laser beams may hit a given marker.
We therefore want to be able to match and use even single marker points.
To that end, \cite{Hong2017} proposed to associate a probabilistic sensor model with each individual point.
We achieve a similar behavior by synthesizing additional points on a circle with radius $r$ around each detected marker point.
We obtained good results with values of $r = \SI{5}{\centi\meter}$.
An example NDT representation of a local map is shown in Fig.~\ref{fig:NdtLayers}.

\subsection{NDT Optimization Function}
Classical NDT scan matching tries to find the best transformation $\bm T$ that transforms the currently measured points $\bm p_i \forall i \in \bm P$ into the coordinate frame of the reference points while maximizing the following NDT score:
\begin{equation}
    \text{score} = \sum_{i}{\exp\left(-\frac{1}{2}(\bm T \bm p_i - \bm q_i)^T \bm\Sigma_i^{-1} (\bm T \bm p_i - \bm q_i)\right)} \label{eq:ndtscore}
\end{equation}
Here, each $\bm p_i$ is one of the currently measured points, while $\bm q_i$ and $\Sigma_i$ are the mean and covariance of the associated NDT cell.

With reflector markers, we simply optimize the weighted sum of associations for both regular points $\bm p_i$ and marker points $\bm p_j$:
\begin{equation}
\begin{split}   
\text{score} = \sum_{i}{\exp\left(-\frac{1}{2}(\bm T \bm p_i - \bm q_i)^T \bm\Sigma_i^{-1} (\bm T \bm p_i - \bm q_i)\right)} \\
+ \sum_{j}{w \exp\left(-\frac{1}{2}(\bm T \bm p_j - \bm q_j)^T \bm\Sigma_j^{-1} (\bm T \bm p_j - \bm q_j)\right)} 
\end{split}
\end{equation}
Where $w$ denotes how much more weight to put on each marker point compared to each regular point.

\subsection{Informed Keyframe Selection Heuristics}
Reflector markers are often not consistently detected in each scan -- especially if they are far away and span only few (down to one) points of the laser scan.
It is thus important to make an informed decision about which scans to use as new keyframe, i.e., which to add to the local map:
In unfortunate cases, the local map may contain only scans in which a rather unreliable marker was never detected, so it would never be considered in scan matching.
In addition to common heuristics about when to choose a new keyframe such as the distance traveled or the quality of scan matching result, we found the ratio of (both regular and reflector) points in the current scan that were successfully associated with an NDT cell to be a good indicator:
If less than a certain ratio (e.g. $70\%$) of all points were successfully associated, the current scan contains significant information about the environment that is not yet represented in the local map.
It should thus be a chosen as a keyframe and added to the local map.

\section{NDT TRACKING APPROACH}
Our second approach is to explicitly track individual markers and to add a special term in the cost function based on these tracks.

\subsection{Marker Tracking}
\newcommand{\track}{\vect}
\newcommand{\detection}{\vecd}

\newcommand{\paramfixedcostdetection}{c_D}
\newcommand{\paramfixedcosttrack}{c_T}
\newcommand{\parammindetections}{n}
\newcommand{\paramwtracks}{w_\text{tracks}}

To use information from reflector markers inside the scan matching algorithm, we first propose a tracking approach that creates continuous tracks from the individual marker detection.

For this purpose, we assume that at each time step, a set of detections $D=\{\detection_1, \dots, \detection_{N_D} \}$ and a set of tracks $T= \{\track_1, \dots, \track_{N_T}\}$ are given.
In the first time step, we initialize $T \gets \emptyset$.
Now, detections and tracks are assigned to each other such that the sum of squared distances between tracks and detections is minimized.
In the assignment algorithm, each detection and each track can be assigned to at most one counterpart.
Unassigned tracks and detections are penalized with fixed costs $\paramfixedcosttrack > 0$ and $\paramfixedcostdetection > 0$, respectively.
In our experiments, we use $\paramfixedcosttrack = \paramfixedcostdetection = (\SI{0.05}{\meter})^2$, i.e. detections are only assigned to tracks closer than this distance.
We refer to the set of assignments as $A = \{(i,j) | \detection_i \text{ is assigned to } \track_j \}$. Then, the cost function is given by
\[ \text{cost}(A) = \sum_{(i,j) \in A} (\detection_j -\track_j)^2 + (N_D - |A|)\cdot \paramfixedcostdetection + (N_T - |A|)\cdot \paramfixedcosttrack \ . \]
This assignment problem can be solved efficiently using the Hungarian algorithm~\cite{Munkres1957} or the Jonker--Volgenant method~\cite{Jonker1987}.

Subsequently, we create a new track for every unassigned detection $\detection_i \in D, \forall\, j (i,j) \notin A $. Each new track is given a track id to uniquely identify it. Unassigned tracks $\track_j \in T, \forall i\, (i,j) \notin A $ are kept at first. They are only discarded if they remain unassigned across a given number of subsequent laser scans.

To improve robustness against brief misdetections, (e.g., misdetections that only occur in a single laser scan), we propose to only use tracks for the scan matching that have been observed at least $\parammindetections \in \mathbb{N}$ times, e.g., $\parammindetections=3$.

After each laser scan, a motion model could be used to predict the track positions in the next laser scan. Since we do not have wheel odometry or an IMU available, in general, our only choice would be to use a motion estimate based on the laser scan matching. However, this might lead to feedback effects where a poor motion estimate might lead to losing or incorrectly assigning the tracks, which in turn would lead to an even worse lidar motion estimate in the next time step. Thus, we only use an identity motion model, which assumes constant positions. This usually works well, as long as the scan rate of the lidar is high enough and the speed of the vehicle is not too great.

\subsection{Scan Matching}
When using reflector markers, the input for the scan matcher consists of a set of all currently measured points $P$, as well as the tracks inside the current scan $T_{A,M} = \{\track_{A,M_1}, \dots, \track_{A,M_k} \}$ and the tracks in the reference scan $T_{R,M} = \{\track_{R,N_1} ,\dots, \track_{R,N_l} \}$ with track ids $M_1,\dots M_k$ and $N_1, \dots, N_l$, respectively. Based on these track ids, we can identify which tracks appear in both scans and define a track-based cost function
\begin{align}
	\text{cost}_\text{tracks} = \sum_{i \in \{M_1, \dots, M_k\} \cap \{ N_1, \dots, N_l\} } \left(\bm T \cdot \track_{A,i} - \track_{r,i} \right)^2 \ .
	\label{eq:costtrackcs}
\end{align}
Based on this cost function, we redefine the NDT score as
\begin{align}
	\text{score}_\text{total} = \text{score}_\text{NDT} - \paramwtracks \cdot \text{cost}_\text{tracks} \ , \label{eq:scoretotal}
\end{align}
where $\text{score}_\text{NDT}$ is given in \eqref{eq:ndtscore} and $\paramwtracks$ is a parameter to adjust the weight of the track-based alignment compared to the geometry-based alignment. The derivatives of the new cost function \eqref{eq:scoretotal} can be computed analytically, which allows for efficient optimization.

Optionally, it is also possible to extend \eqref{eq:costtrackcs} to include an error term for the orientation of each track if a detector is used that provides the marker normals (such as the wall detector proposed in Sec.~\ref{sec:detection}).

\subsection{Informed Keyframe Selection Heuristics}
Similar to the NDT layer approach, we propose to extend the keyframe selection criteria of the lidar odometry algorithm to take reflector markers into account. In particular, we create a new keyframe every time a new track is detected. This way, new tracks will be immediately available in the local map for future scan matches.

Since the local map is composed of multiple scans, a track may occur more than once in the local map, i.e., there are multiple tracks with the same id. To resolve this, there are several options:
\begin{itemize}
	\item Use the oldest track to reduce drift.
	\item Use the newest track to increase robustness against markers that are not entirely static.
	\item Use the mean of all tracks with the same id to be more resilient in the presence of noisy measurements.
\end{itemize}
We currently use the oldest track in our implementation.

\section{EXPERIMENTS}
\subsection{Data Recording}
\begin{figure}[ht]
	\centering
	\includegraphics[height=0.35\textwidth]{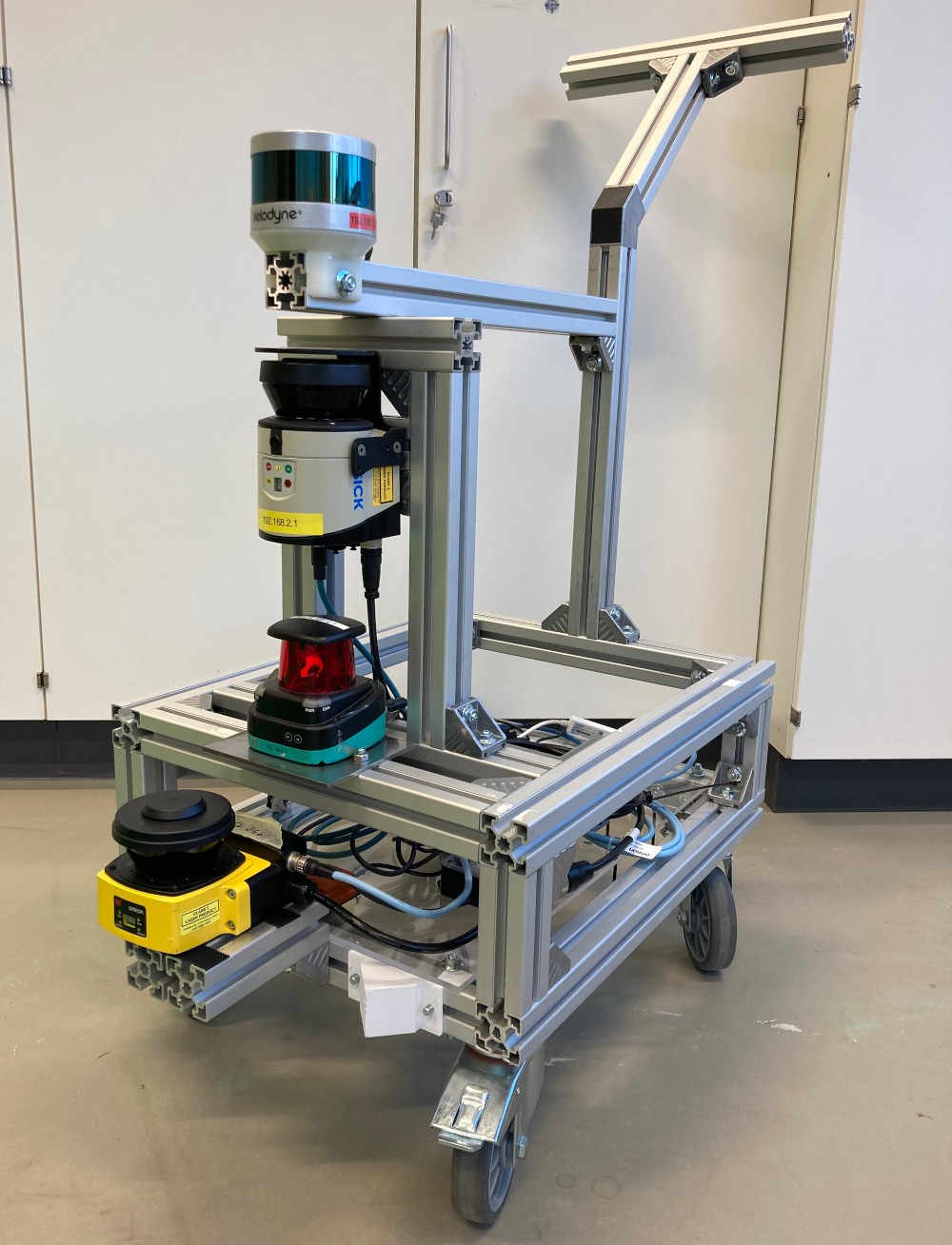}
	\caption{The cart used to record data with multiple laser scanners at once.}
	\label{fig:Waegele}
\end{figure}

We built the cart shown in Fig.~\ref{fig:Waegele} to conveniently record data with multiple laser scanners at the same time.
The laser scanners from bottom to top are:
Omron OS32C, Pepperl \& Fuchs R2000, SICK LMS151, and Velodyne VLP-16.
The lower three are 2D sensors often used for localization in intralogistics environments.
The VLP-16 is a 3D sensor that, combined with a 3D-variant of our SLAM software, acts as a reference system, since it always measured enough geometric features to not get lost.
We also use an Xsens MTi inertial measurement unit, which is used to correct small rotations in roll and pitch and for motion prediction and undistortion within the duration of each laser scan.

We used this recording cart in three different locations, which were either known to be difficult for 2D laser localization systems or where we assumed that problems would occur due to potential false positive marker detections:
Two different production plants with both featureless corridors and many reflective metal parts and an industrial outdoor storage area for large metal parts with ramps and featureless corridors.
We ended up with 24 (R2000), 20 (LMS151), and 14 (OS32C) recordings per scanner: The OS32C was unusable in the outdoor storage area due to its too short measurement range and some LMS151 recordings were unusable due to networking issues while recording.

\subsection{Detector Evaluation}
\subsubsection{Efficient Labeling}
Evaluating the reflector marker detector requires labeled data. Labeling individual scans is both tedious and difficult or even impossible to do reliably (cf. Fig.~\ref{fig:compare_lidars}).
We instead label markers by marking their bounding boxes within the accumulated point cloud of a generated SLAM map\footnote{using \url{https://github.com/Earthwings/annotate}}.
This is much easier, since we know where we placed the markers in the environment.
Since the estimated pose of each scan is known after SLAM, we can transform points and bounding boxes from the map to each scan's frame and vice versa.
We perform this annotation for at least one SLAM map from each environment where we recorded data and each 2D laser scanner considered here.

\subsubsection{Evaluation}
For each detected marker, we check whether it lies within the bounding box of a labeled marker to decide whether to treat it as a true or false positive.
We grow the bounding boxes by a small distance (\SI{10}{\centi\meter}) to accommodate small inaccuracies in the estimated scan pose.
We count missed detections by first testing all labeled bounding boxes for their visibility in a scan:
If there is a point in the laser scan that lies within the bounding box and there was no marker detection within the box, we count this as a false negative.
We are not interested in the potentially large number of true negatives.

\begin{figure}[ht!]
	\centering
	\includegraphics[width=0.9\columnwidth]{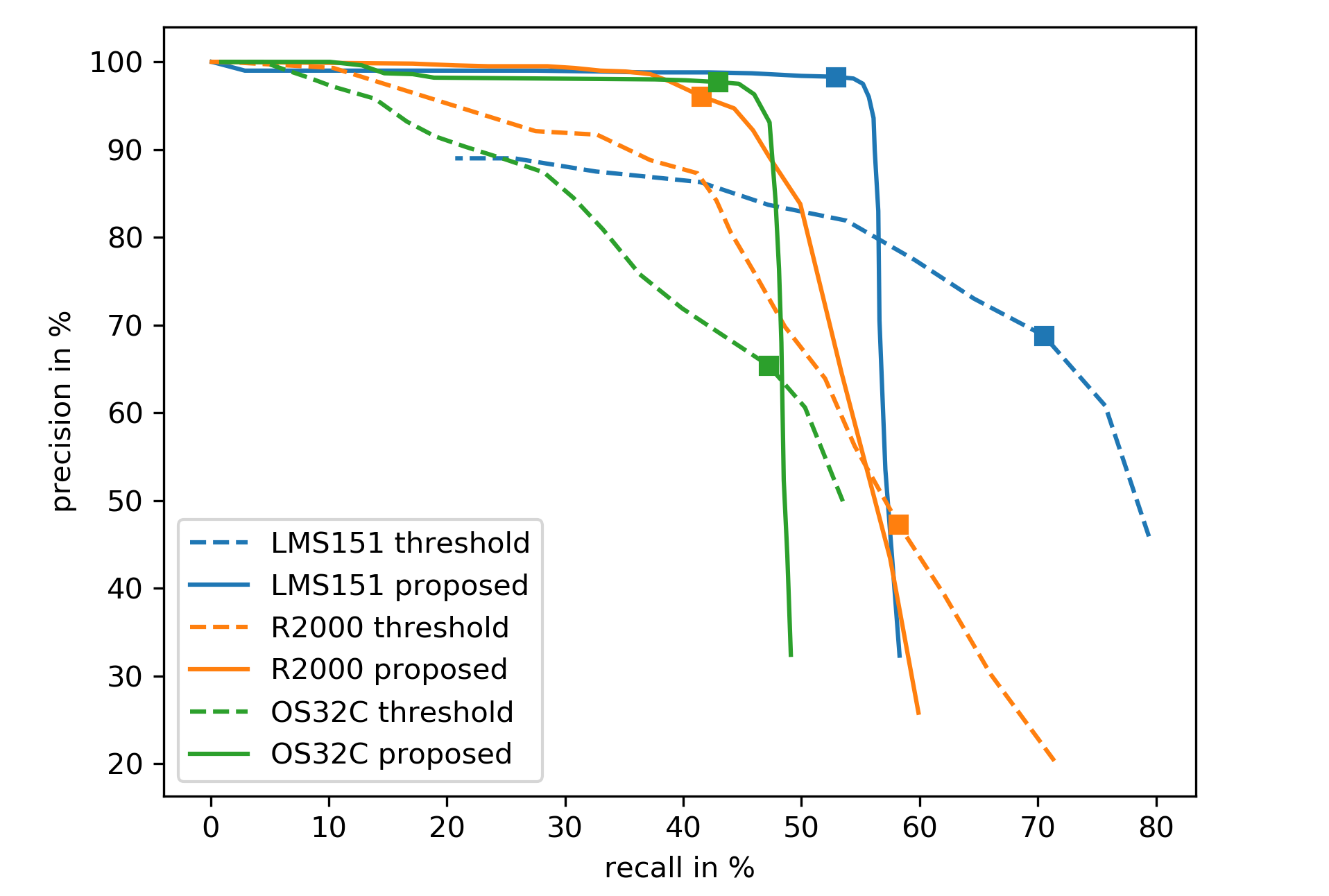}
	\caption{Precision-recall curves for the evaluated detectors and sensors.
	Small squares denote the operating point (using one per-sensor intensity threshold) chosen for further SLAM experiments.}
	\label{fig:precisionRecallCurve}
\end{figure}

We evaluate the \textit{proposed} detector in Section~\ref{sec:detection} and vary its intensity threshold parameter to plot its performance as a precision-recall curve in Fig.~\ref{fig:precisionRecallCurve}. The small squares denote the operating point for the intensity values we deemed most suitable for usage in SLAM, since they offer the best compromise between the required high precision and a still usable recall ($1000$ for LMS151, $500$ for R2000, and $8000$ for OS32C, see Table~\ref{tab:lidars}).

In addition, we also show the precision-recall curve for a baseline \textit{threshold} detector, which classifies each point above the intensity threshold as a marker without any further checks.
Patches of consecutive marker points are combined to a single detection located at their mean position. 
We choose the same intensity threshold parameters as for the \textit{proposed} detector above, which leads to the operating points marked by small squares.

\begin{figure}[ht!]
	\centering
	\includegraphics[width=0.98\columnwidth]{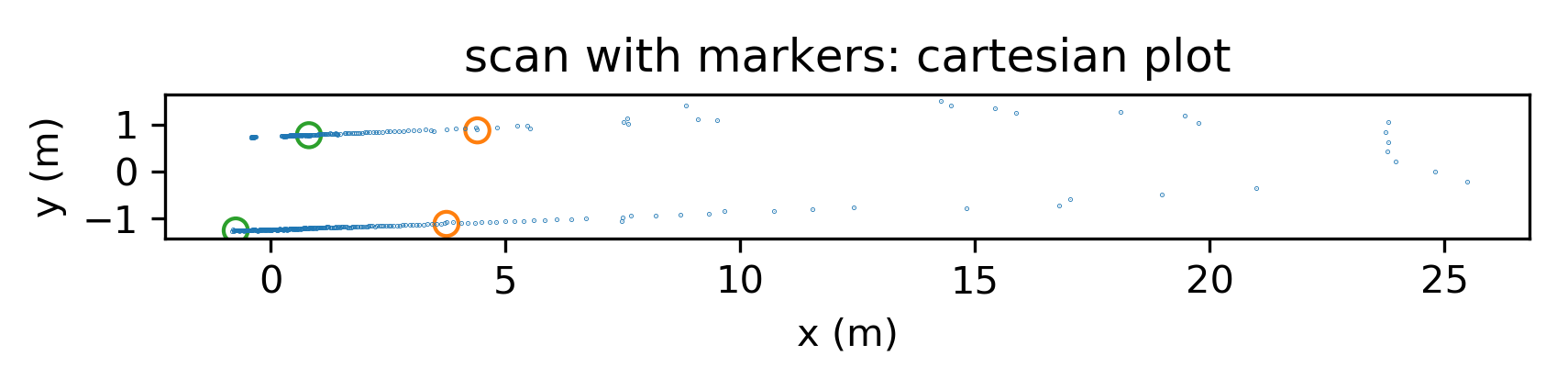}
	\includegraphics[width=0.98\columnwidth]{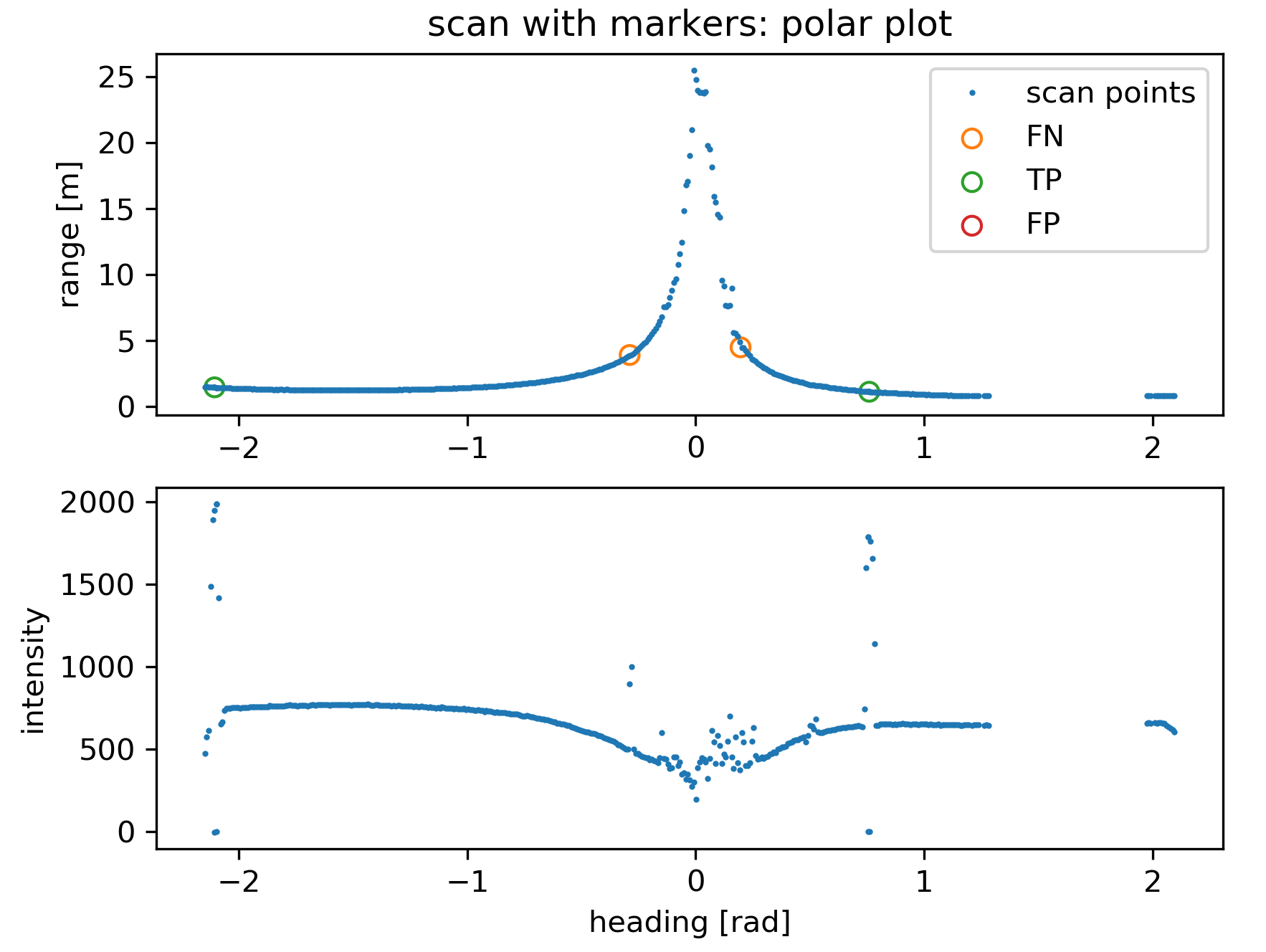}
	\caption{Example of (missed) detections in a narrow corridor.}
	\label{fig:detectionResultsExample}
\end{figure}

The recall rates obtained using the \textit{threshold} detector at the operating point can be interpreted as a hard upper bound:
Detections missed by it do not lead to a single point above the intensity threshold, which means we cannot hope to detect these with our proposed detector.
In our experience, these stem mainly from markers seen at steep angles, which are unfortunately quite common in long and narrow corridors.
An example scan in such a situation is shown in Fig.~\ref{fig:detectionResultsExample}, where the peaks in intensity for the two markers further away in the corridor are either rather small or non-existent.

Our proposed detector improves over the baseline \textit{threshold} detector by filtering out most false positives to increase its precision, while trying not to disregard too many true positives.
The decrease in recall is therefore expected to a certain extent.
We investigated the remaining false positives of our \textit{proposed} detector and found that they are mainly caused by other reflective objects found in the original environments.
This is not a problem long as these objects do not move.
Their unintentional contribution as reflector markers may even help with scan matching.

\subsection{SLAM Evaluation}
We evaluate the SLAM performance by comparing the resulting trajectory after the entire recording was processed, and after the whole SLAM graph was finally optimized by comparing it to a reference trajectory.
This reference trajectory is generated in the same way, but using the 3D variant of our pipeline that uses data from the 3D laser scanner.
We are mainly interested in identifying and counting cases where SLAM failed, which usually manifests in the trajectory estimate either getting stuck even though it should be moving or by moving too much or in the wrong direction. We identify these cases by computing the absolute tracking error (ATE)~\cite{Zhang2018} between both trajectories after correcting for the fact that the scanners are not mounted at exactly the same pose on the recording cart and aligning both SLAM trajectories to the same reference frame by solving the absolute orientation problem \cite{Eggert1997}.
The failure cases can automatically be identified by checking whether their maximum ATE is above \SI{1.0}{\meter}, in which case we consider the SLAM run to have failed.

An overview over the resulting success rates are shown in Table~\ref{tab:SLAMSuccessRates}.
The numbers of recordings are slightly different for each sensor since not all sensors were available for all recordings.
The success rates without markers are quite different to begin with due to differences in range, angular resolution, and measurement rate for the evaluated sensors.
\begin{table}
	\begin{tabular}{l|c|cc|cc}
		\toprule
		sensor & \multirow{2}{1cm}{without \\ markers} & \multicolumn{2}{c}{NDT layers} & \multicolumn{2}{c}{NDT tracking}  \\
		& & thresh. & prop. & thresh. & prop. \\
		\midrule
		SICK LMS151 & 66.7\% & 70.0\% & 95.0\% & 45.0\% & 100.0\% \\
		\midrule
		P\&F R2000 & 80.0\% & 83.3\% & 95.8\% & 41.7\% & 91.7\% \\
		\midrule
		Omron OS32C & 21.4\% & 64.3\% & 100.0\% & 42.9\% & 92.9\% \\
		\bottomrule
	\end{tabular}
	\caption{SLAM success rates for different laser scanners and methods to integrate reflector markers.}
	\label{tab:SLAMSuccessRates}
\end{table}

Even if absolute percentages are similar, they do not show whether the same runs fail or succeed in each case. 
To that end, we collected Table~\ref{tab:SLAMConfusionTable} which contains information similar to a confusion matrix, showing how many runs that failed without markers succeeded with markers and vice versa.

\begin{table}
	\begin{tabular}{p{1cm}|p{1cm}|p{1cm}|p{0.7cm}|p{0.7cm}|p{0.7cm}|p{0.7cm}}
		\rot{detector} & \rot{sensor} & \rot{marker integration} & \rot{succeeds only w/ markers} & \rot{succ. only w/o markers} & \rot{same result} \\
		\midrule
		\multirow{6}{1cm}{threshold}
		& \multirow{2}{1cm}{LMS151}
		& layer & 25.0\% & 20.0\% & 55.0\% \\
	   & & tracking & 20.0\% & 40.0\% & 40.0\% \\
		& \multirow{2}{1cm}{R2000}
		& layer & 8.3\% & 4.2\% & 87.5\% \\
	   & & tracking & 4.2\% & 41.7\% & 54.2\% \\
		& \multirow{2}{1cm}{OS32C}
		& layer & 50.0\% & 7.1\% & 42.9\% \\
	   & & tracking & 42.9\% & 21.4\% & 35.7\% \\
	   \midrule
	   \multirow{6}{1cm}{proposed}
		& \multirow{2}{1cm}{LMS151}
		& layer & 30.0\% & 0.0\% & 70.0\% \\
	   & & tracking & 35.0\% & 0.0\% & 65.0\% \\
		& \multirow{2}{1cm}{R2000}
		& layer & 16.7\% & 0.0\% & 83.3\% \\
	   & & tracking & 12.5\% & 0.0\% & 87.5\% \\
		& \multirow{2}{1cm}{OS32C}
		& layer & 78.6\% & 0.0\% & 21.4\% \\
	   & & tracking & 71.4\% & 0.0\% & 28.6\% \\
		\bottomrule
	\end{tabular}
	\caption{Confusion table for SLAM results}
	\label{tab:SLAMConfusionTable}
\end{table}

We can see that from Tables~\ref{tab:SLAMSuccessRates} and \ref{tab:SLAMConfusionTable} that using reflector markers in SLAM in combination with a too simple \textit{threshold} detector with too low precision does not help but in fact often makes it worse.

In combination with the more precise \textit{proposed} detector, however, using reflector markers and either of the two presented approaches clearly helps to increase robustness of SLAM:
Many recordings in which SLAM failed before can now be solved accurately, whereas none of the runs that succeeds without markers fails with markers enabled.

In our experiments the NDT layer approach has a slight edge over the NDT tracking approach.
We believe this is related to markers often being detected unreliably, which may prevent these from being tracked.
Being able to match detected markers even when they were not consistently tracked seems to help the NDT layer approach more than it hinders it.
We believe the situation could be different for a different detector or the same detector at a different operating point with lower precision and higher recall, in which case the NDT tracking approach could take advantage of its capability to filter out incorrect matches that were not tracked.

But even with reflector markers enabled, we do not reach a success rate of \SI{100}{\percent} for our recorded dataset in all combinations.
This is because we intentionally placed markers sparsely for some recordings to test the limits of our system. 
As shown in Figure~\ref{fig:detectionResultsExample}, detecting markers seen at steep angles can be surprisingly challenging, which can lead to situations in which no markers are in view anymore. This should certainly be kept in mind in future applications.

\section{CONCLUSIONS \& FUTURE WORK}
In this paper, we proposed and evaluated approaches to make a lidar-based SLAM system more robust by using reflector markers, which can easily be installed in difficult environments.
We described a detector that processes laser scans and reliably detects wall-mounted reflective markers of a known size.
Furthermore, we described two different methods to take advantage of detected markers during scan matching, specifically in NDT matching.
Finally, we evaluated both methods and showed that they succeed in significantly increasing the robustness of lidar-based SLAM.

We saw that marker detection within our system is of little use if the user did not place enough markers or if they are too far from each other.
How densely the markers should be placed depends on the distance and angle at which they can still be detected, which varies for different lidar sensors.
Future work looking into sensor-specific recommendations for placement and guiding the user in this task may be required.

In this work, we only focused on using markers for SLAM, but the proposed approach also applies to lidar-based localization. It may be worthwhile to look into other types of markers and also using markers for global relocalization as in~\cite{MeyerDelius2011}. It may also be useful to dynamically enable and disable the usage of markers to achieve optimal performance within a given environment.

\bibliographystyle{IEEEtran}
\bibliography{../bib/gkbosch}

\begin{thebibliography}{10}
\providecommand{\url}[1]{#1}
\csname url@samestyle\endcsname
\providecommand{\newblock}{\relax}
\providecommand{\bibinfo}[2]{#2}
\providecommand{\BIBentrySTDinterwordspacing}{\spaceskip=0pt\relax}
\providecommand{\BIBentryALTinterwordstretchfactor}{4}
\providecommand{\BIBentryALTinterwordspacing}{\spaceskip=\fontdimen2\font plus
\BIBentryALTinterwordstretchfactor\fontdimen3\font minus
  \fontdimen4\font\relax}
\providecommand{\BIBforeignlanguage}[2]{{%
\expandafter\ifx\csname l@#1\endcsname\relax
\typeout{** WARNING: IEEEtran.bst: No hyphenation pattern has been}%
\typeout{** loaded for the language `#1'. Using the pattern for}%
\typeout{** the default language instead.}%
\else
\language=\csname l@#1\endcsname
\fi
#2}}
\providecommand{\BIBdecl}{\relax}
\BIBdecl

\bibitem{Guivant2000}
J.~Guivant, E.~Nebot, and S.~Baiker, ``Localization and map building using
  laser range sensors in outdoor applications,'' \emph{Journal of Robotic
  Systems}, vol.~17, no.~10, pp. 565--583, 2000.

\bibitem{Ghallabi2019}
F.~Ghallabi, M.-A. Mittet, G.~El-Haj-Shhade, and F.~Nashashibi, ``{LIDAR}-based
  high reflective landmarks ({HRL})s for vehicle localization in an {HD} map,''
  in \emph{2019 {IEEE} Intelligent Transportation Systems Conference
  ({ITSC})}.\hskip 1em plus 0.5em minus 0.4em\relax {IEEE}, oct 2019.

\bibitem{Wang2021}
S.~Wang, X.~Chen, G.~Ding, Y.~Li, W.~Xu, Q.~Zhao, Y.~Gong, and Q.~Song, ``A
  lightweight localization strategy for {LiDAR}-guided autonomous robots with
  artificial landmarks,'' \emph{Sensors}, vol.~21, no.~13, p. 4479, jun 2021.

\bibitem{Davis2019}
S.~Davis, K.~G. Ricks, and R.~A. Taylor, ``Reflective fiducials for
  localization with 3d light detection and ranging scanners,'' \emph{{IEEE}
  Access}, vol.~7, pp. 45\,291--45\,300, 2019.

\bibitem{Besl1992}
P.~J. {Besl} and N.~D. {McKay}, ``A method for registration of {3-D} shapes,''
  \emph{IEEE Transactions on Pattern Analysis and Machine Intelligence},
  vol.~14, no.~2, pp. 239--256, Feb. 1992.

\bibitem{Biber2003}
P.~Biber and W.~Strasser, ``The normal distributions transform: a new approach
  to laser scan matching,'' in \emph{Proceedings 2003 {IEEE}/{RSJ}
  International Conference on Intelligent Robots and Systems ({IROS} 2003)
  (Cat. No.03CH37453)}.\hskip 1em plus 0.5em minus 0.4em\relax {IEEE}, 2003.

\bibitem{Olson2009}
E.~B. Olson, ``Real-time correlative scan matching,'' in \emph{2009 IEEE
  International Conference on Robotics and Automation}, 2009, pp. 4387--4393.

\bibitem{Grisetti2010a}
G.~{Grisetti}, R.~{Kümmerle}, C.~{Stachniss}, and W.~{Burgard}, ``A tutorial
  on graph-based {SLAM},'' \emph{IEEE Intelligent Transportation Systems
  Magazine}, vol.~2, no.~4, pp. 31--43, 2010.

\bibitem{MunozSalinas2020}
R.~Muñoz-Salinas and R.~Medina-Carnicer, ``Ucoslam: Simultaneous localization
  and mapping by fusion of keypoints and squared planar markers,''
  \emph{Pattern Recognition}, vol. 101, p. 107193, 2020.

\bibitem{Zaganidis2018}
A.~Zaganidis, L.~Sun, T.~Duckett, and G.~Cielniak, ``Integrating deep semantic
  segmentation into 3-d point cloud registration,'' \emph{IEEE Robotics and
  Automation Letters}, vol.~3, no.~4, pp. 2942--2949, 2018.

\bibitem{Zaganidis2017}
A.~Zaganidis, M.~Magnusson, T.~Duckett, and G.~Cielniak, ``Semantic-assisted 3d
  normal distributions transform for scan registration in environments with
  limited structure,'' in \emph{2017 IEEE/RSJ International Conference on
  Intelligent Robots and Systems (IROS)}.\hskip 1em plus 0.5em minus
  0.4em\relax IEEE, 2017, pp. 4064--4069.

\bibitem{Nuechter2005}
A.~N{\"u}chter, O.~Wulf, K.~Lingemann, J.~Hertzberg, B.~Wagner, and H.~Surmann,
  ``3d mapping with semantic knowledge,'' in \emph{RoboCup 2005: Robot Soccer
  World Cup IX}, A.~Bredenfeld, A.~Jacoff, I.~Noda, and Y.~Takahashi,
  Eds.\hskip 1em plus 0.5em minus 0.4em\relax Berlin, Heidelberg: Springer
  Berlin Heidelberg, 2006, pp. 335--346.

\bibitem{Kuemmerle2011}
R.~Kümmerle, G.~Grisetti, H.~Strasdat, K.~Konolige, and W.~Burgard,
  ``g\textsuperscript{2}o: A general framework for graph optimization,'' in
  \emph{Proc. IEEE Int. Conf. Robotics and Automation}, May 2011, pp.
  3607--3613.

\bibitem{Andreasson2017EgoMotionUncertainty}
H.~Andreasson, D.~Adolfsson, T.~Stoyanov, M.~Magnusson, and A.~J. Lilienthal,
  ``Incorporating ego-motion uncertainty estimates in range data
  registration,'' in \emph{2017 IEEE/RSJ International Conference on
  Intelligent Robots and Systems (IROS)}, 2017, pp. 1389--1395.

\bibitem{Hong2017}
H.~{Hong} and B.~H. {Lee}, ``Probabilistic normal distributions transform
  representation for accurate {3D} point cloud registration,'' in \emph{Proc.
  IEEE/RSJ Int. Conf. Intelligent Robots and Systems (IROS)}, Sep. 2017, pp.
  3333--3338.

\bibitem{Munkres1957}
J.~Munkres, ``Algorithms for the assignment and transportation problems,''
  \emph{Journal of the Society for Industrial and Applied Mathematics}, vol.~5,
  no.~1, pp. 32--38, Mar. 1957.

\bibitem{Jonker1987}
R.~Jonker and A.~Volgenant, ``A shortest augmenting path algorithm for dense
  and sparse linear assignment problems,'' \emph{Computing}, vol.~38, no.~4,
  pp. 325--340, dec 1987.

\bibitem{Zhang2018}
Z.~Zhang and D.~Scaramuzza, ``A tutorial on quantitative trajectory evaluation
  for visual(-inertial) odometry,'' in \emph{2018 {IEEE}/{RSJ} International
  Conference on Intelligent Robots and Systems ({IROS})}.\hskip 1em plus 0.5em
  minus 0.4em\relax {IEEE}, oct 2018.

\bibitem{Eggert1997}
D.~W. Eggert, A.~Lorusso, and R.~B. Fisher, ``Estimating 3-d rigid body
  transformations: a comparison of four major algorithms,'' \emph{Machine
  Vision and Applications}, vol.~9, no.~5, pp. 272--290, Mar 1997.

\bibitem{MeyerDelius2011}
D.~Meyer-Delius, M.~Beinhofer, A.~Kleiner, and W.~Burgard, ``Using artificial
  landmarks to reduce the ambiguity in the environment of a mobile robot,'' in
  \emph{2011 {IEEE} International Conference on Robotics and Automation}.\hskip
  1em plus 0.5em minus 0.4em\relax {IEEE}, may 2011.

\end{thebibliography}

\end{document}